\documentclass[letterpaper]{article} 
\usepackage{aaai25}  
\usepackage{times}  
\usepackage{helvet}  
\usepackage{courier}  
\usepackage[hyphens]{url}  
\usepackage{graphicx} 
\usepackage{natbib}  
\usepackage{caption} 
\usepackage{multirow}
\title{HandDiffuse: Generative Controllers for Two-Hand Interactions via Diffusion Models}
\author {
    Pei Lin\textsuperscript{\rm 1},
    Sihang Xu\textsuperscript{\rm 1},
    Hongdi yang\textsuperscript{\rm 1},
    Yiran Liu\textsuperscript{\rm 1},
    Xin Chen\textsuperscript{\rm 2},
    Jingya Wang\textsuperscript{\rm 1},
    Jingyi Yu\textsuperscript{\rm 1},
    Lan Xu\textsuperscript{\rm 1}    
}
\affiliations {
    \textsuperscript{\rm 1}ShanghaiTech University, School of Information Science and Technology, Shanghai, China\\
    \textsuperscript{\rm 2}PCG, Tencent, China\\
    }

\begin{document}

\maketitle

\begin{abstract}
 Existing hands datasets are largely short-range and the interaction is weak due to the self-occlusion and self-similarity of hands, which can not yet fit the need for interacting hands motion generation. To rescue the data scarcity, we propose HandDiffuse12.5M, a novel and real dataset that consists of temporal sequences with strong two-hand interactions. HandDiffuse12.5M has the largest scale and richest interactions among the existing two-hand datasets. We further present a strong baseline method HandDiffuse for the controllable motion generation of interacting hands using various controllers. Specifically, we apply the diffusion model as the backbone and design two motion representations for different controllers. To reduce artifacts, we also propose Interaction Loss which explicitly quantifies the dynamic interaction process. Our HandDiffuse enables various applications, i.e., motion in-betweening and trajectory controled generation. Experiments show that our method outperforms the state-of-the-art techniques in motion generation.
The vivid two-hand motions generated by our method can also construct synthetic datasets and enhances the accuracy of existing hand motion capture algorithms.
\end{abstract}

%
\begin{links}
    \link{Datasets}{https://handdiffuse.github.io/}
\end{links}


\section{Introduction}\label{sec:intro}
The tightly interacting hands play a crucial role in human emotional expression and communication, serving as a vital component for physical interaction with oneself. Controllable hand motion generation not only enhances the experience of augmented reality/virtual reality (AR/VR) but also help robotics and avatars to express their emotion in another dimension.

So far, lots of works have emerged in the field of human body motion generation~\cite{mdm2022human,zhang2022motiondiffuse,chen2023executing,jiang2023motiongpt}, but the generation of temporal hands motions in strong interaction remains blank as the lack of strong interacting hands datasets. However, the dataset acquisition is time-consuming and expensive due to the interacting hands present self-occlusion, self-similarity, and complex articulations.

There are only a few datasets~\cite{Tzionas:IJCV:2016,Moon_2020_ECCV_InterHand2.6M,zuo2023reconstructing,li2023renderih,moon2023reinterhand} focusing on hands with strong interactions. Most of them, i,e, InterHand2.6M~\cite{Moon_2020_ECCV_InterHand2.6M}, Two-hand 500K~\cite{zuo2023reconstructing} and RenderIH~\cite{li2023renderih}, provides large-scale discrete frames but the available temporal sequences remain sparse. The severe lack of sufficient temporal motions makes them unsuitable for the motion generation task of two-hand interactions. Only recently, the concurrent work Re:InterHand~\cite{moon2023reinterhand} provides large-scale interaction data, yet it focuses on the motion capture of interacted hands under diverse lighting conditions.

\begin{figure}[t]
  \centering
   \includegraphics[width=1\linewidth]{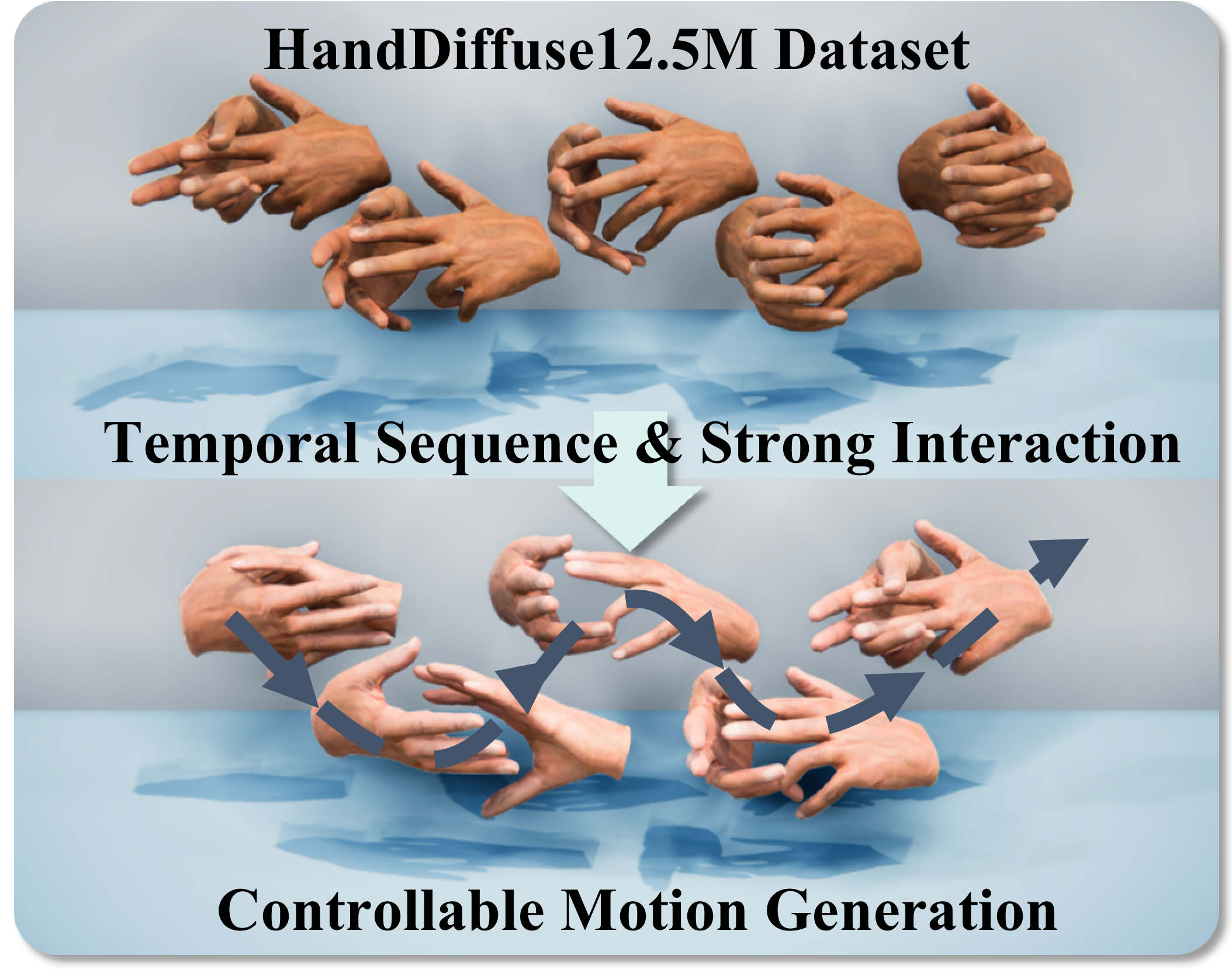}
\caption{\textbf{Overview:} The proposed HandDiffuse12.5M benchmark dataset consists of temporal sequences with strong interaction. Based on it, we propose HandDiffuse, a strong baseline for the motion generation of interacting hands.}
  \label{fig:teaser}
\end{figure}

To break the data scarcity, in this paper, we first contribute a real dataset, \textit{HandDiffuse12.5M}, which focuses on capturing diverse temporal sequences with strong two-hand interactions. Our dataset consists of various modalities, including synchronous 50-view videos, 3D key points, and hand poses. It covers interacting hand motions in random and complex poses, daily communications, finger dances, and Chinese daily sign language with 250K temporal frames($\approx$ 2 hours), resulting in roughly 12.5M pictures. To address the issue of occlusion, which is unavoidable in hand interactions, we manually annotate the key points in 240K images. Note that our HandDiffuse12.5M dataset owns the largest scale not only in the duration but also in the number of pictures among the existing available interacting hands datasets. Thus, it not only opens up the research for temporally consistent two-hand motion generation but also brings huge potential for future exploration in hand modeling.

Based on our novel dataset, we further provide a strong baseline algorithm named \textit{HandDiffuse}, which tailors the diffusion model~\cite{ho2020denoising} to generate vivid interacting hands motions under diverse explicit control.
Note that hand motion is an abstract form of emotional expression and ten fingers always gather together during interactions. Therefore, it is difficult to describe the motions with concrete text prompts and the control for motion generation needs to be explicit. 
To this end, we adopt several explicit controllers as the conditions for our diffusion models, including the discrete keyframes, the trajectories of wrists and 2D keypoints.
These conditionings enable various downstream applications, ranging from hand motion in-betweening to trajectory-aided hand motion completion which connects hand motions with body motions (see Figure\ref{fig:teaser}), and can also assist in hand motion reconstruction.
In HandDiffuse, we propose Interacting Hands DDIM which is inspired by modular design  and come up with two specific motion representations for different controllers, as well as a novel Interaction Loss for the dynamic process of hand interaction, which helps to reduce the artifacts. 
Finally, we provide a thorough evaluation of our approach as well as a comparison with state-of-the-art motion generation methods~\cite{mdm2022human,liang2023intergen,lee2024interhandgen,mueller2023buddi}. We also provide results to show that HandDiffuse can contribute to data augmentation for other datasets. Our preliminary outcomes indicate that controllable hand motion generation remains a challenging problem, and our dataset will consistently benefit further investigations of this new research direction. 
To summarize, our main contributions include:
\begin{itemize}
    \item We contribute a large-scale real dataset named HandDiffuse12.5M, which provides accurate and temporally consistent tracking of human hands under diverse and strong two-hand interactions. 
    
    \item We are the first to specifically target on hand-only interaction motion generation and we propose a strong baseline for generating interacting hand motions from various explicit controllers, enabling vivid motion in-betweening, trajectory conditioning and hands reconstruction.

    \item We introduce Interacting Hands DDIM, two motion representations as well as an Interaction Loss to significantly improve the generation results of two-hand interactions. The generated motions can contribute to data augmentation for other datasets. 
    
    \item Our dataset, related codes will be made publicly available to stimulate further research. 
\end{itemize}


\section{Related Works}
\label{sec:related}

\begin{table*}[ht] 
\setlength{\tabcolsep}{1mm}
	\centering
		\begin{tabular}{lccccc}
			\hline
			Dataset   & Temporal Sequence & Source & \# Views & Manually Annotation & Appearance \\
			\hline
RenderIH~\cite{li2023renderih}           & No        & Synthesis      & -   & No  &  Natural\\
Two-hand 500K~\cite{zuo2023reconstructing} & Partial   & IMU+Synthesis  & -   & No  &  -      \\
InterHand2.6M~\cite{Moon_2020_ECCV_InterHand2.6M} & Partial   & RGB            & 80  & Yes &  Lab       \\
Re:InterHand~\cite{moon2023reinterhand}    & Yes       & RGB            & 26  & No  &  Relighting\\
HandDiffuse12.5M(Ours)        & Yes       & RGB            & 50  & Yes &  Lab\\
			\hline
		\end{tabular}
 	    \caption{\textbf{Dataset Comparisons.} We compare our HandDiffuse12.5M dataset with other existing publicly available interacting-hand datasets. Our HandDiffuse12.5M dataset contains both strong interaction and temporal sequences, and has the largest scale.}
    \label{tab: dataset comparison}
\end{table*}

\paragraph{\textbf{Hand Dataset.}}
Although hand datasets have significantly accelerated the development of VR/AR, embodied AI and human-object interaction. 
Obtaining ground truth for specific dual-hand interaction motion from real-world captured data is challenging~\cite{Moon_2020_ECCV_InterHand2.6M,li2023renderih}. Moon et al.~\cite{Moon_2020_ECCV_InterHand2.6M} have attempted to combine manual and automated annotations, but manual annotations are labor-intensive. Recently, Moon et al.~\cite{moon2023reinterhand} have improved the detection accuracy of 2D keypoints and reconstructed hands to facilitate fully automated annotation schemes, but these methods are still time-consuming.

As a result, some works have established synthetic hand datasets to enhance annotation precision~\cite{zuo2023reconstructing,li2023renderih,zb2017hand,Lin_2021_WACV,gao2022dart}. Among them, RenderIH~\cite{li2023renderih} increases the diversity of data through re-lighting, while DART~\cite{gao2022dart} provides a wide range of hand accessories to bridge the gap between synthetic and real data. However, synthetic datasets fail to balance the diversity of motion and the continuity of sequences. Some works~\cite{li2023renderih,zuo2023reconstructing} sample single-frame actions for the left and right hands separately from real datasets and solve the occlusion problem between the hands through optimization. However, such optimization methods are not effective for temporal sequences. Therefore, very few papers~\cite{Tzionas:IJCV:2016,Moon_2020_ECCV_InterHand2.6M,moon2023reinterhand} have collected temporal interacting hand datasets~\cite{li2023renderih}.

\paragraph{\textbf{Hands Capture \& Reconstruction.}}
The reconstruction of both hands is a significant challenge in the area of human motion capture~\cite{zuo2023reconstructing}. 
The release of the InterHand2.6M~\cite{Moon_2020_ECCV_InterHand2.6M} has motivated many regression-based methods~\cite{rong2021ihmr,Zhang2021twohand,9880324,Hampali_2022_CVPR_Kypt_Trans,di2022lwahand,fan2021digit,9710274,meng2022hdr,moon2023interwild}. These regression-based methods significantly boost the development of VTuber and VR/AR devices due to their balance between real-time performance and accuracy. Some works~\cite{9880324,park2023extractandadaptation} proposed a Transformer-based network with the cross-attention between right and left hands. Recently, Zuo et al.~\cite{zuo2023reconstructing} constructed a learning-based prior to capture dual-hands which achieved great results. Building on the impressive performance of diffusion models in image generation~\cite{ronneberger2015unet,song2020denoising,ho2020denoising}, DiffHand~\cite{li2023diffhand} combines hand reconstruction with diffusion models.  The existing hands capture algorithms can also get enhanced by synthetic data. RenderIH\cite{li2023renderih} and DART\cite{gao2022dart} have shown that the synthetic hand datasets can improve the capture accuracy.

\paragraph{\textbf{Human Motion generation.}}
In recent years, there has been a growing interest in the field of human motion generation, with many studies exploring the generation of human motions based on conditioning signals. These conditioning signals are used to guide the generation of specific types of motions, such as motion in-betweening~\cite{MotionInfilling, Harvey_2020, qin2022motion}, trajectory control~\cite{Zhang_2021_ICCV, Rempe_2023_CVPR, karunratanakul2023gmd, xie2023omnicontrol} and unconstrained motion synthesis~\cite{Pavlakos_2019_CVPR, yan2019convolutional, Zhao_2020_CVPR}. Leading by the pioneering work of MDM~\cite{tevet2023human}, human motion generation models based on diffusion architecture~\cite{liang2023intergen,zhang2022egobody, du2023avatars} have been shown to gain better motion diversity and expressiveness compared to prior works based on GAN~\cite{lin2018human} or VAE~\cite{cai2021unified, petrovich2021action}. 
However, most existing works in this area have primarily focused on generating whole-body motions, only few works\cite{lee2024interhandgen,zhang24both} try to generate hands' motion in recent.
Utilizing the capabilities of the diffusion models, we propose a novel method HandDiffuse to enable interacting hands motion generation tasks with control signals such as key frames and trajectory.


\section{HandDiffuse12.5M Dataset}\label{sec:dataset}

\paragraph{\textbf{Capture system}}   
Existing hand motion capture solutions include IMU gloves, elastic sensor gloves, marker-based systems, and multi-view camera capture. However, due to the severe occlusion in hand interactions, marker-based systems are not suitable; IMU gloves and elastic sensor-based capture solutions theoretically address the issue of visual occlusion, but the IMU gloves are strict to calibration and it is easy to accidentally touch the sensor during the hands interaction; the elastic sensor-based capture solutions lacks the restoration of all degrees of freedom for finger joints.
Therefore, we have opted for a multi-view RGB approach combined with manual annotation, similar to InterHand2.6M. Our capture system consists of 50 mounted video cameras recording interaction sequences at a frame rate of 30-60 frames per second (fps). We adjust the resolution to 3840 × 2160 to allocate more pixels for capturing hand movements. 
 We design four categories of hand interaction motions for recording: random but complex hand interactions, daily communication, finger tutting dance, and sign language. Each motion sequence is controlled to last approximately one minute limited by the hardware. More detailed description about the capture system has been shown in Appendix.

\begin{figure}[t]
  \centering
   \includegraphics[width=1\linewidth]{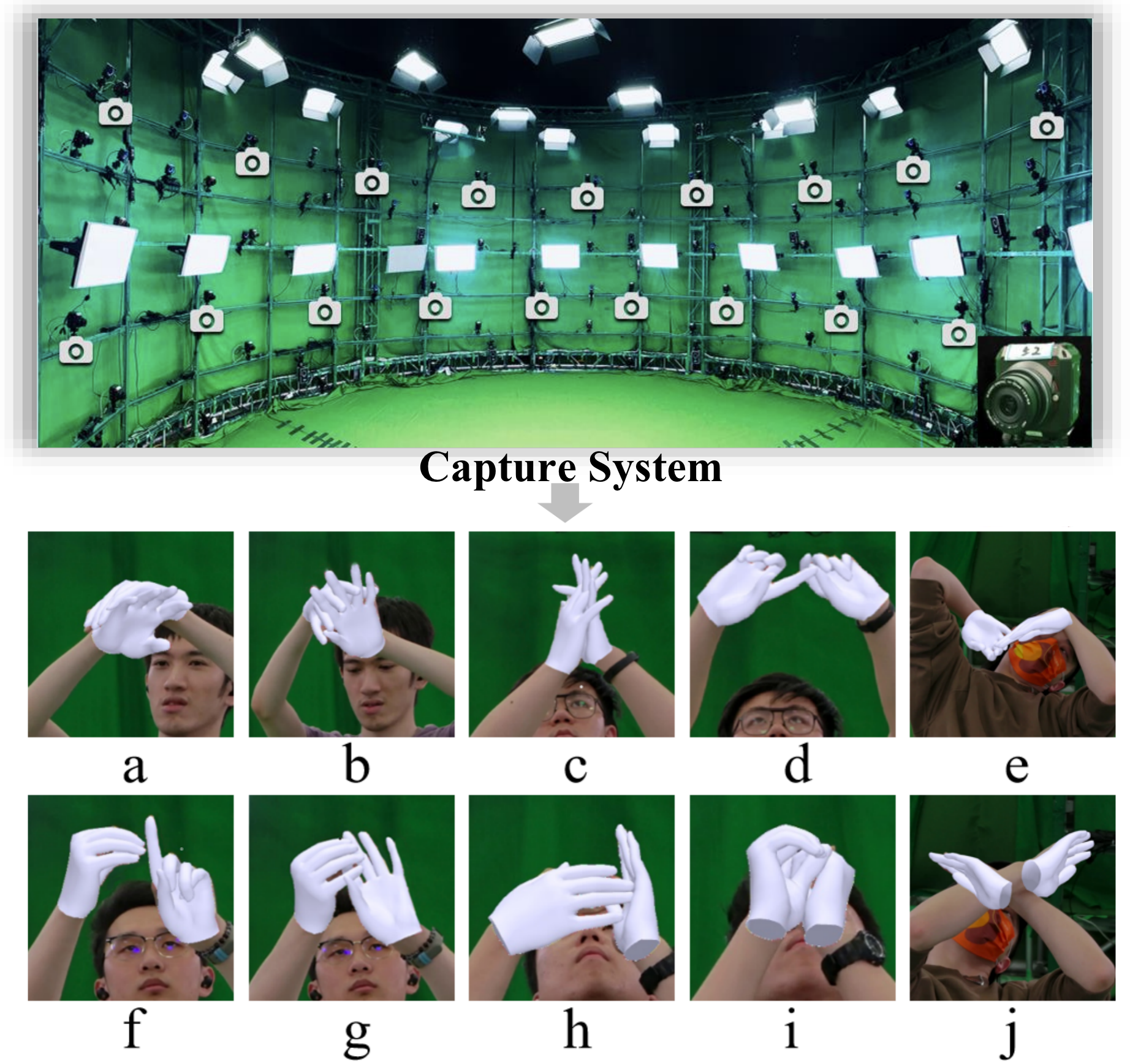}
\caption{The capture system and reprojectoin of MANO. The proposed HandDiffuse12.5M benchmark dataset consists of strong and various interaction with accurate annotation.}
  \label{fig:dataset}
\end{figure}

\paragraph{\textbf{2/3D joint coordinates \& MANO fitting}}    
Upon securing multi-view video sequences, we utilize DWPose\cite{dwpose} which is  the SOTA 2D keypoints detector to extract the hands 2D keypoints.
The subsequent triangulation of these 2D keypoints yields 3D keypoints in world coordinate, which are then integrated into 3D pose by optimizing MANO \cite{MANO:SIGGRAPHASIA:2017} which is differentiable. 
The frames with failed fitting will be selected out and manually re-annotated with 2D key points on the images. 
Figure\ref{fig:dataset} has shown some MANO reprojection results which contains strong and various hands interaction. Due to space limitation, a more detailed procedure for annotating the dataset and the approach we used to fit MANO is illustrated in the Appendix.


\begin{figure}[t]
  \centering
   \includegraphics[width=1\linewidth]{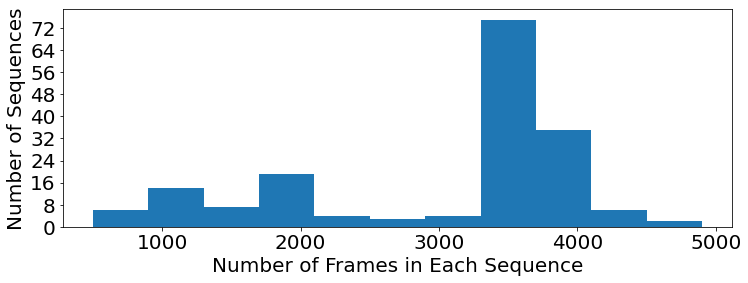}
\caption{The distribution of HandDiffuse12.5M’s temporal frames.}
  \label{fig:dataset count}
\end{figure}

\paragraph{\textbf{Quantitative Evaluation for our Dataset.}}  
Our HandDiffuse12.5M dataset focuses on temporal sequences which has more than 250K temporal frames. This makes it the largest dataset among all of the interacting hands datasets as presented in Table \ref{tab: dataset comparison}. As shown in Figure \ref{fig:dataset count}, each sequence has around 1000 to 4000 temporal frames and the average is 2955 temporal frames per motion sequence, with the median of 3630 frames.

\begin{figure}[t]
  \centering
   \includegraphics[width=1\linewidth]{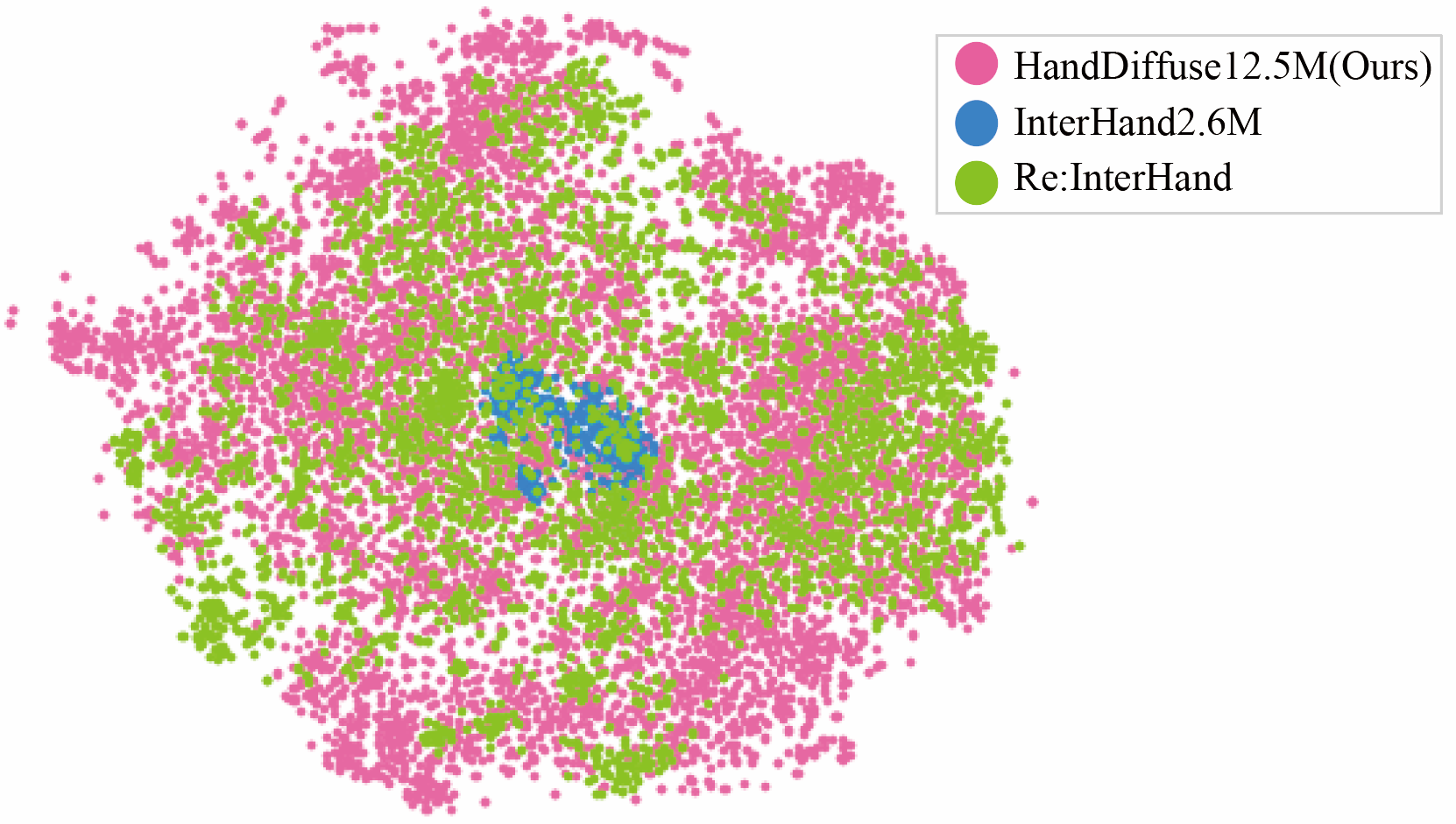}
\caption{t-SNE visualization of HandDiffuse12.5M(Ours), InterHand2.6M and Re:InterHand. For InterHand2.6M, we only choose its temporal frames. The result indicates the diversity and richness of HandDiffuse12.5M.}
  \label{fig:dataset tsne}
\end{figure}

Our dataset’s average shortest distance between two hands' joints is 3.61cm, superior to InterHand2.6M’s 4.04 cm and Re:InterHand’s 3.84 cm. 
We classify a sample as ``contacting'' if the shortest distance between two hands’ vertices is under 3 mm. With a contacting hands ratio of 57.3\%, our dataset exceeds InterHand2.6M’s 44.74\% and Re:InterHand’s 52.7\%. 
We randomly selected 5000 images from the dataset to manually annotate the 2D keypoints. The error between the annotation results and the reprojection of 3D keypoints is 11.3 pixels per keypoint in a 3840×2160 image space. The MANO fitting error is 4.3mm per joint.

We also compare our dataset with InterHand2.6M \cite{Moon_2020_ECCV_InterHand2.6M} and Re:InterHand \cite{moon2023reinterhand}, both of which also contain hand interaction motion sequences, by visulizing the t-SNE results shown in  Figure \ref{fig:dataset tsne}.
Figure \ref{fig:dataset tsne} illustrates that our dataset exhibits a greater variety of interaction poses compared to InterHand2.6M and Re:InterHand. This is attributed to our dataset's explicit focus on hand interaction sequences for generation tasks.

In summary, HandDiffuse12.5M comprises 12.5 million images and 250K temporal frames, making it the largest two-hand dataset to date. It provides a variety of strongly interactive motion and accurate annotations.


\section{Method}\label{sec:method}
\begin{figure*}[t] 
    \centering
		\includegraphics[width=\linewidth]{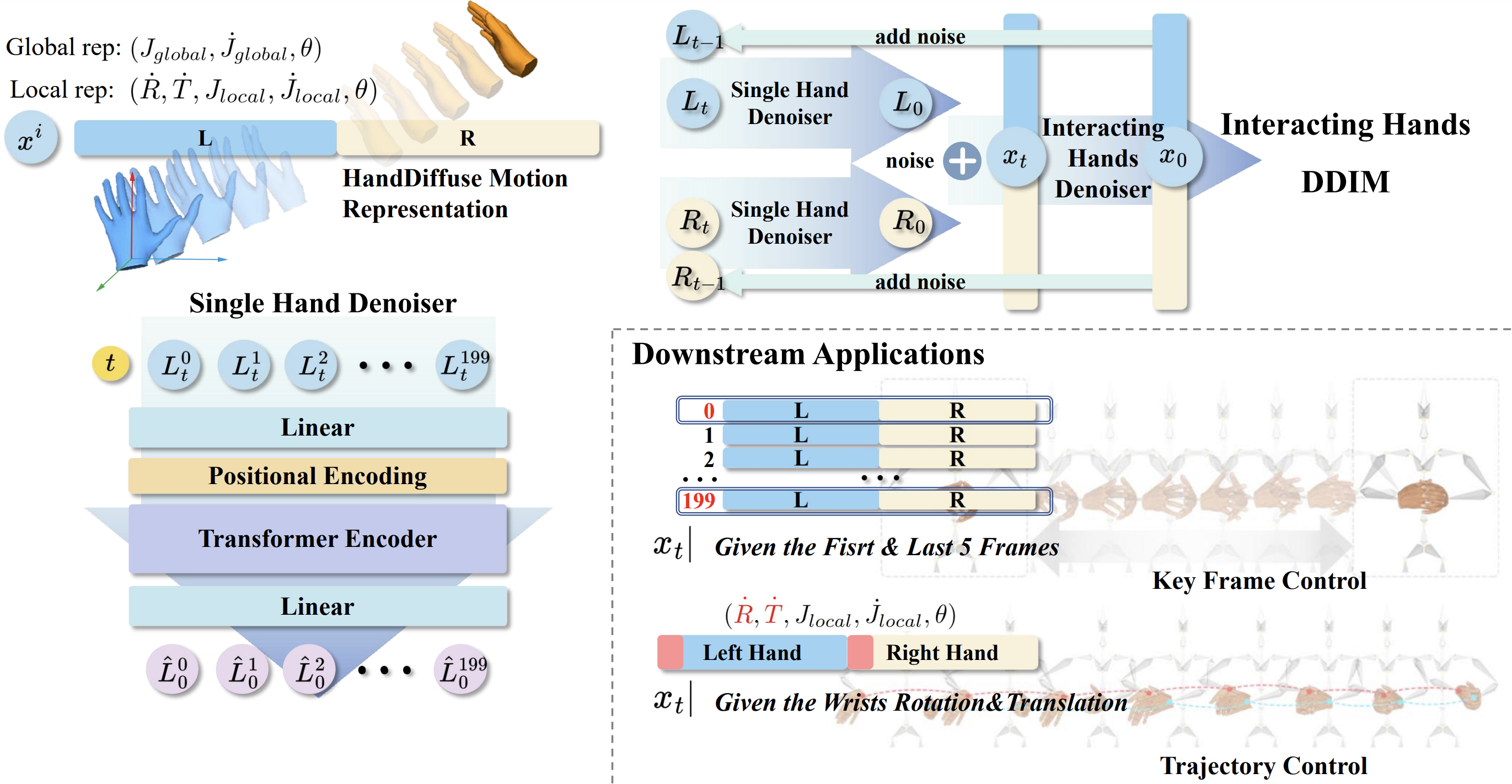} 
	\caption{\textbf{(Left) Overview of HandDiffuse} We first generate the motions for each hand separately by training Single Hand Denoiser which only focuses on local poses. The generated two single hands' local pose are concatenated with global information(noise) and transformed into designed motion representation. The interacting hands denoiser further optimizes the interaction process. 
     \textbf{(Right) Overview of Downstream Applications.}
     When the control is key frames, we generate the final motions $\mathbf{\hat{x}_0^{1:N}}$ by giving the first 5 frames $\mathbf{x_{GT}^{:5}}$ and the last 5 frames $\mathbf{x_{GT}^{-5:}}$ before denoising. 
     When the control os wrists' trajectories, we generate the final motion $\mathbf{\hat{x}_0}$ by giving the root angular velocity and linear velocity in each frame before denoising. Other controls like 2D keypoints has been illustrated in Appendix due to the space limitation.} 
	\label{fig: framework pipeline} 
\end{figure*} 

An overview of HandDiffuse is shown in Figure \ref{fig: framework pipeline}. With HandDiffuse12.5M dataset, the goal is to synthesize two hands motion $\mathbf{x^{1:N}}$ without or with designed controllers $c$. 
These selected controllers are common in body motion generation and are also able to dictate the synthesis of interacting hands motions, such as key frames (motion in-betweening), trajectory of both wrists (trajectory control) and 2D keypoints(3D hands reconstruction).
In contrast to single-person human motion generation, the task of generating interacting hands motion requires the network to learn not only the local motion of each hand but also the dynamic process of global interaction.   
Therefore, we adopts a modular design to learn the local pose of single hand and dynamic interacting process separately. We also propose two specific hand motion representations for different controllers and Interaction Loss ($\mathbf{Loss}_{interaction}$) to model this dynamic interacting process better. 

Inspired by recent work, we adopt the diffusion model as the backbone which has shown remarkable generative ability in image generation and human motion generation. Diffuse is modeled as a Markov noising process and the ground truth $\mathbf{x_0^{1:N}}$ is diffused to $\mathbf{x_t^{1:N}}$ by adding t-step independent Gaussian noise $ \epsilon \sim \mathcal{N}(0,I)$. This process can be formulated as 
$$q(\mathbf{x_t|x_0}) = \sqrt{\overline{\alpha}_t}\mathbf{x_{0}}+(1-\overline{\alpha}_t),$$
where $\overline{\alpha}_t = \prod_{s=0}^t \alpha_s$ and $\alpha_t = 1-\beta_t$. $\beta_t$ is the cosine noise variance schedule and $q(\mathbf{x_t})$ is near $\mathcal{N}(0,I)$ when t is big enough. Then, $\mathbf{x_t}$ is sent to a denoiser $\mathcal{G}$ conditioned on timestep $t$ and different conditions $c$ to predict $\mathbf{\hat{x}_0}$:
$$\mathbf{\hat{x}_0} = \mathcal{G}(\mathbf{x_t} , t, c).$$

\paragraph{\textbf{Two Denoisers \& Interacting Hands DDIM}}
HandDiffuse adopts a modular design and applies two denoisers: Single Hand Denoiser(\textbf{SHDe}) and Interacting Hands Denoiser(\textbf{IHDe}) to focus on the different parts of the whole interacting process as shown in Figure \ref{fig: framework pipeline}.

SHDe is used to learn the characteristics of local hand motion, so we set the global translation and global rotation of the hand to 0, and extract the local joints' positions ($ J_{local}, \dot{J}_{local}$) and pose($\theta $) as input. By mirroring the right hand to the left hand, we only trained one SHDe. By minimizing the
\begin{equation}
    \mathbf{Loss}_{reconSH} = \Vert L_{GT} - \hat{L}_0 \Vert_2, \label{1}
\end{equation}
where $\hat{L}_0$ is the denoised single hand parameter and $ L_{GT}$ is the ground truth, we model the local motion of single hand.

After training the SHDe and freezing its parameters, we concatenate the generated local motions of two hands with the global information which is t-step noise and transform the whole vector to designed motion representation.
IHDe focuses on the global information and will also finetune the local motion. 
During the training stage, after completing a round of denoising, we minimize
\begin{equation}
    \mathbf{Loss}_{IHDe} = \mathbf{Loss}_{reconDH} + \mathbf{Loss}_{interaction},  \label{2}
\end{equation}
\begin{equation}
    \mathbf{Loss}_{reconDH} = \Vert x_{GT} - \hat{x}_0 \Vert_2, \label{3}
\end{equation}
where $\hat{x}_0 $ is the denoised two hands parameter and $ x_{GT}$ is the ground truth, and $\mathbf{Loss}_{interaction}$ will be illustrated later. In the inference stage, after denoising, we re-add the noise of $t-1$ steps. We refer to this process as Interacting Hands DDIM. Subsequent evaluations have proven the excellence of the modular design.

In terms of model architecture, we adopted an encoder-only transformer similar to MDM\cite{mdm2022human}. SHDe and IHDe have the same structure but different dimensions.

\paragraph{\textbf{Motion Representations for Interacting Hands.}}
\label{sec: Motion representation and parameterization}

For different tasks, we have proposed two specific motion representations, which we refer to as local representation (local rep) and global representation (global rep). 

The local rep is inspired by HumanML3D\cite{Guo_2022_CVPR_humanml3d}, a widely used approach in human motion generation. In each frame, the pose of left hand is defined by a tuple of $(\dot{R}, \dot{T}, J_{local}, \dot{J}_{local}, \theta)$  and the right hand is 
$( R_{init},  T_{init},\dot{R}, \dot{T}, J_{local}, \dot{J}_{local}, \theta )$, 
where $\dot{R} \in \mathrm{R}^4$ and $\dot{T} \in \mathrm{R}^3$ represent root angular velocity in quaternion and linear velocity respectively. $J_{local} \in \mathrm{R}^{3j}$ and $\dot{J}_{local} \in \mathrm{R}^{3j}$ denote the local joint positions and velocities in root space, with $j=21$ represents the number of joints. $\theta \in \mathrm{R}^{4i}$ represents the local joint rotations in quaternion representation, following the skeleton structure of MANO with $i=15$. Each sequence is aligned with the left wrist of the first frame as the coordinate system origin. Both left and right hands are rotated to align the left-hand $R$ vector with the direction (1, 0, 0) which is shown in \ref{fig: framework pipeline}. Therefore, the right hand has $R_{init}\in \mathrm{R}^{4}$ and $T_{init}\in \mathrm{R}^{3}$.

The global rep is defined as a tuple of ($J_{global}$, $\dot{J}_{global}$, $\theta$) for both hands, where $J_{global}$ represents the positions of hand joints in the world coordinate system and two hands are normalized by employing the same approach as local rep.

During the experimental process, we observed that the local rep performed better in the task where the global rotation and translation of the hand are given. The global rep achieved better results in the unconditional generation, the control of key frames(motion in-betweening).

\begin{figure*}[t] 
    \centering
		\includegraphics[width=\linewidth]{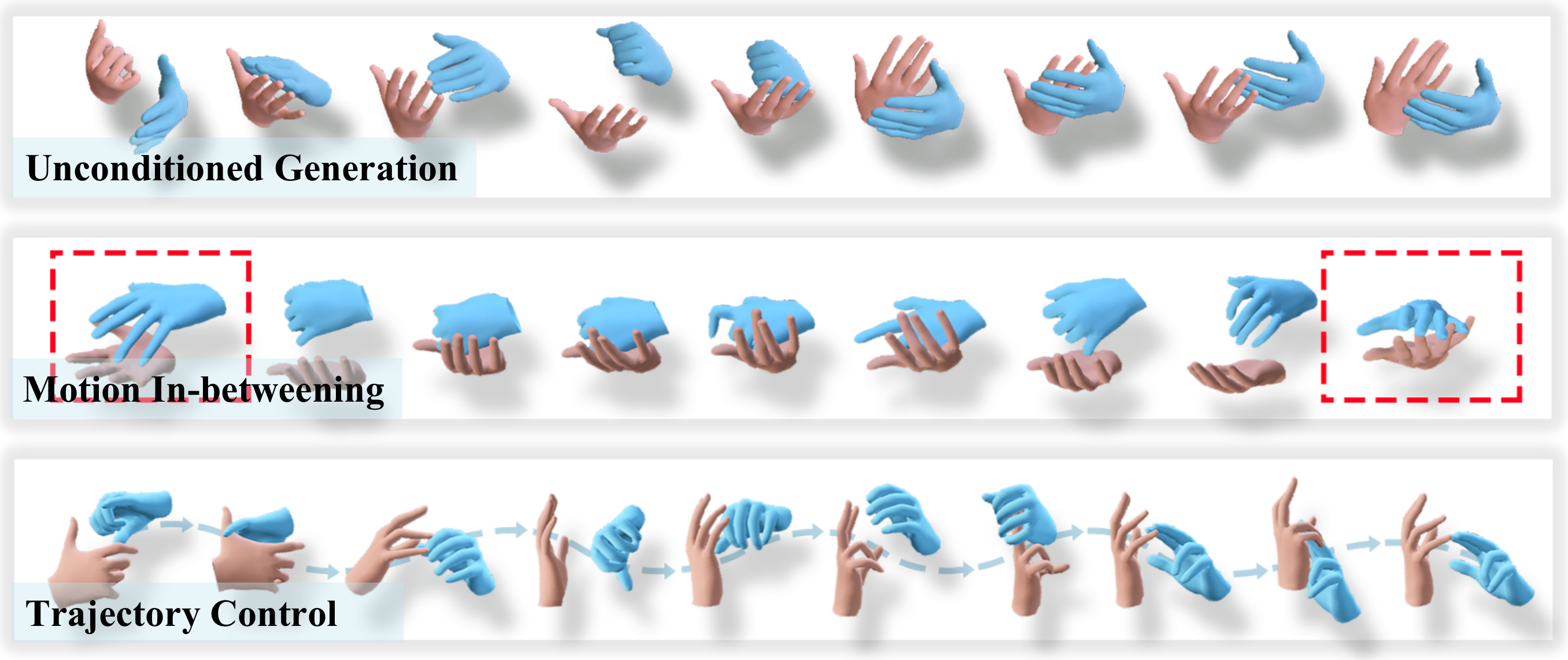} 
	\caption{\textbf{Two Downstream Applications of HandDiffuse.} The generation results of downstream applications of HandDiffuse.} 
	\label{fig: Gallery} 
\end{figure*}

\paragraph{\textbf{Interaction Loss \ \ $\mathbf{Loss}_{interaction}$}}
\label{Interaction Loss}

The first part of the $\mathbf{Loss}_{interaction}$ is the  contact potential loss($\mathbf{Loss}_{potential}$ ). In each frame, we first calculate a $25\times 25$ distance matrix $\mathbf{D}$ between the 21 left-hand joints and 21 right-hand joints with 4 more joints sampled in each palm. Since the interaction between the hands involves fine-grained movements, we should focus on situations where the finger joints are very close to each other. Inspired by the concept of Contact Potential Field\cite{yang2021cpf}, we further represent the distances using the elastic potential energy of a spring model and it can be described as
\begin{equation}
\mathbf{P} = 1/2 \mathbf{K} \ relu(\mathbf{\tau}-\Vert \mathbf{D}_{pred}\Vert_2)^2  , \label{4}
\end{equation}
 where $\mathbf{K}$ represents the coefficient of elasticity and $\mathbf{\tau}$ denotes the distance threshold. The closer the distance between the finger joints, the larger the potential energy obtained, and the distance will be filtered out when it is over the threshold. 
 
 Moreover, by considering the direction of the distance, we can avoid ambiguity caused by penetration, and the final contact potential loss is described as
\begin{equation}
\mathbf{Loss}_{potential} = |\mathbf{P}_{pred}-\mathbf{P}_{gt}|(1+\overrightarrow{\mathbf{D}_{pred}} \times \overrightarrow{\mathbf{D}_{gt}}),  \label{5}
\end{equation}
where $(\times)$ means cross product. More is shown in Appendix.

\begin{figure*}[t] 
    \centering
		\includegraphics[width=\linewidth]{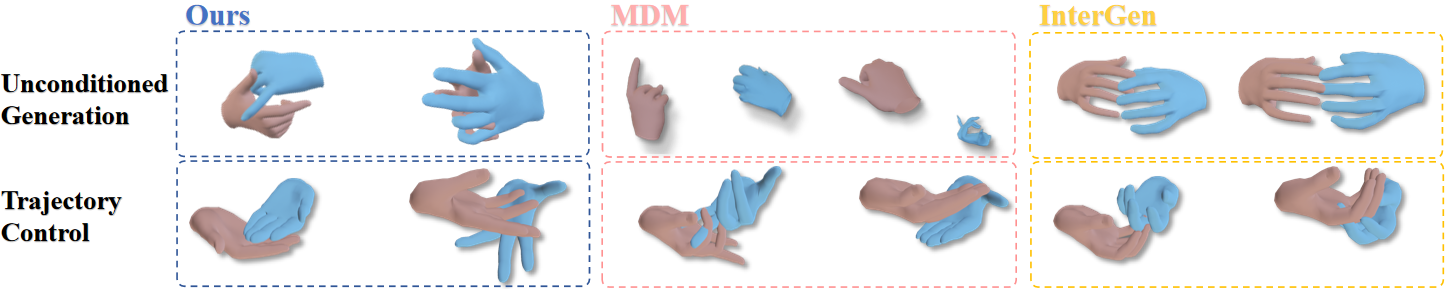} 
	\caption{\textbf{Qualitative comparison for different methods. } For Unconstrained Generation, MDM tend to generate motion with hands shifting away from each other, while InterGen would generate motions with a mean pose. In terms of Trajectory Control, MDM yields satisfactory results, whereas InterGen generates motion without incorporating interaction.} 
	\label{fig: Qualitative Comparison for different methods} 
\end{figure*}

We observed that MANO has strict requirements for the bone length. Therefore, the second part of the $\mathbf{Loss}_{interaction}$ is the Shape Loss($\mathbf{Loss}_{shape}$ ). We ensure the rationality of hand shape by calculating the differences in bone lengths between the left and right hands, as well as the differences between the predicted shape and the ground truth (GT). This is described as follows:
\begin{equation}
    \mathbf{Loss}_{shape} = |\mathbf{BL}_{left}-\mathbf{BL}_{right}| + |\mathbf{BL}_{pred}-\mathbf{BL}_{GT}|, \label{6}
\end{equation}
where $BL$  denotes the bone length.

\paragraph{\textbf{Downstream Applications}}\label{sec: application}
For the task of key frames control,  we provide the ground truth for the previous five frames $\mathbf{x_{GT}^{:5}}$ and the last five frames $\mathbf{x_{GT}^{-5:}}$ for the diffused motion $\mathbf{x_t^{1:N}}$. For the task of trajectory control, the goal is to generate the reasonable local motion given the global rotation and translation of both hands. In this case, we provide $\dot{R}^{1:N}$ and $\dot{T}^{1:N}$ as control signal for the diffused motion $\mathbf{x_t^{1:N}}$. The results are presented in Figure \ref{fig: Gallery}.

We also explore HandDiffuse by applying the control of 2D keypoints, and the experiments have shown that HandDiffuse can also solve the task of hands reconstruction. Due to the space limitation, the results are presented in Appendix.


\section{Experiment}\label{sec: experiment}
In this section, we conduct several experiments to evaluate our method HandDiffuse for the task of interacting hands motion generation. We particularly evaluate 
(1) the comparison of our method against previous state-of-the-art approaches, and 
(2) the ablation study. The above evaluations are conducted specifically on the two controllers (key frames and trajectory) and other controller(2D keypoints) is evaluated in Appendix due to the space limitation. 
More experiments on the ability of HandDiffuse to augment existing datasets are presented in Appendix.
The models have been trained with $T = 1000$ diffusing steps and a cosine noise schedule. The denoiser has 8 layers, 4 heads and the feedforward dimension is 1024. We set the length of generated motion $N$ to 200 in all experiments. All of them have been trained on a single \textit{NVIDIA GeForce RTX 3090} GPU for about two days.

\begin{table*}[ht]
\setlength{\tabcolsep}{1mm}
\centering
\begin{tabular}{c|ccccccccc}
\hline
\multirow{2}{*}{Method} & \multicolumn{3}{c|}{Unconditioned} & \multicolumn{3}{c|}{In-betweening}    & \multicolumn{3}{c}{Trajectory Control}  \\
                        & FID$\downarrow$                  & Diversity$\uparrow$            & \multicolumn{1}{c|}{SDF$\downarrow$} & FID$\downarrow$                  & Diversity$\uparrow$            & \multicolumn{1}{c|}{SDF$\downarrow$} & FID$\downarrow$                  & Diversity$\uparrow$            & SDF$\downarrow$                  \\ \hline
real                    &           0.050           &       12.034               &    1.146                      &                      &                      &                          &                      &                      &                      \\
MDM\cite{tevet2023human}           &     0.526   &  10.366                    &  1.74                     & 0.616                         &  10.258                    &    1.621                  &       0.205                   &     11.265                 &       1.572                                \\

InterGen\cite{liang2023intergen}                  &     0.251                 &       4.928               &          \textbf{ 0}               &      \textbf{0.269}                &   5.385                   &   \textbf{1.032}                      &             0.402         &       9.102               &          3.572            \\ 
BUDDI\cite{mueller2023buddi}                  &    0.853                  &        8.282              &               2.293           &       1.211               &           7.341           &      2.482                    &               0.921       &               7.497       &   1.842                   \\
InterHandGen\cite{lee2024interhandgen}                  &     0.437                 &       10.238               &       1.731                   &            0.734          &          8.620            &     1.819                     &         0.539             &           10.124           &        1.282              \\ 
HandDiffuse(ours)                  &       \textbf{0.161}               &          \textbf{11.829}            &   1.553                        &           0.273           &          \textbf{11.554}            &      1.562                    &             \textbf{0.173}         &    \textbf{11.758}                  &          \textbf{1.141}            \\ \hline
\end{tabular}
\caption{\textbf{Quantitative comparisons} of different methods. We compare the \textbf{FID} and \textbf{Diversity} for different methods. '$\downarrow$' indicates that results are better with lower metrics, $\uparrow$ is on the contrary. \textbf{Bold} means the best result.}
\label{tab: method comparison}
\end{table*}

\begin{table*}[!ht]
\setlength{\tabcolsep}{1mm}
\centering
\begin{tabular}{ccccccc}
\hline
Task                     & \multicolumn{2}{c}{Unconditioned} & \multicolumn{2}{c}{In-betweening} & \multicolumn{2}{c}{Trajectory Control} \\
                         & FID$\downarrow$             & SDF$\downarrow$             & FID$\downarrow$             & SDF$\downarrow$             & FID$\downarrow$                & SDF$\downarrow$               \\ \hline
real                     & 0.050           & 1.146           &     -            &  -               &         -           &      -             \\
w Global rep.            & \textbf{0.161}           & \textbf{1.553}           & \textbf{0.273 }          & \textbf{1.562}           & 0.418              & 1.347             \\
w Local rep.             & 0.431           & 1.962           & 0.567           & 1.971           & \textbf{0.173}              & \textbf{1.141}             \\
w/o SHDe                 & 0.168           & 1.575           & 0.282           & 1.571           & 0.182              & 1.538             \\
w/o ${Loss}_{potential}$ & 0.171           & 1.606           & 0.317           & 1.628           & 0.210              & 1.614             \\
w/o ${Loss}_{shape}$     & 0.390           & 1.583           & 0.512           & 1.579           & 0.380              & 1.245             \\ \hline
Complete model           & \textbf{0.161}           & \textbf{1.553}           & \textbf{0.273 }          & \textbf{1.562}           & \textbf{0.173}              & \textbf{1.141}             \\ \hline
\end{tabular}
\caption{\textbf{Quantitative comparisons.} ``w/o SHDe" means IHDe is the only denoiser. \textbf{SDF} is only calculated on frames with penetraion.}
\label{table: ablation study}
\end{table*}

\paragraph{\textbf{Comparisons of different methods}}
We compare our approach with MDM~\cite{tevet2023human}(single human motion generation), InterGen~\cite{liang2023intergen}(interacting human motion generation), BUDDI\cite{mueller2023buddi}(latent representation of interacting human) and InterHandGen\cite{liang2023intergen}(interacting hands motion generation of single frame). 
The metrics we applied are common used Fréchet Inception Distance (FID), Diversity and sdf loss(SDF).
FID can measure the realistic of the generated motion. 
Diversity can prove that the motion do not maintain the mean pose for the sake of preserving realistic. 
SDF would take the positive value while zero otherwise, when the hand joints penetrate with each other. Using the formulation, we can measure the severity of penetration during the whole motion.
The strategy we applied to compute these metrics is inspired by MotionCLIP\cite{tevet2022motionclip} and is illustated in Appendix.
We modify these methods by adapting their motion representation to enable hand motion generation.
Specifically, we adapted InterHandGen to generate continuous frames following its original pipeline in which we first generated 200 frames of left-hand motion, then generated the right-hand motion based on the left hand.
We adapted BUDDI for continuous frame generation and treat the motion of each frame as a token, and retained the "Person Embed" component to distinguish different hands.
As depicted in Figure~\ref{fig: Qualitative Comparison for different methods}, both MDM and InterGen exhibit limitations when generating hand motion. They tend to produce motion sequences with severe joint penetration, unnatural interaction or static mean pose.
The resulst of HandDiffuse has the highest diversity as shown in Table \ref{tab: method comparison}. Because the motion generated by InterGen without condition is static, the FID is lower than us and the penetration does not exist.


\paragraph{\textbf{Ablation Study}}\label{ablation}

At first, we demonstrate the effectiveness of the different motion representations in different tasks in Table \ref{table: ablation study}.
%
 The qualitative result is shown in Appendix. By incorporating the $\mathbf{Loss}_{interaction}$, we mitigate the issue of joint penetration, resulting in improved motion generation quality.  
 Table~\ref{table: ablation study} demonstrates that our Interacting Hands DDIM with modular deisgn  achieves the highest performance when using the proper motion representation and $\mathbf{Loss}_{interaction}$.



\section{Conclusion}
We present HandDiffuse12.5M, the largest interacting hands dataset with strong interaction and temporal sequences, to open up the research direction of interacting hands motion generation. We propose a strong baseline method HandDiffuse which is the first interacting hands motion generation approach along with some useful controllers: key frames control, trajectory control and 2D keypoints control. 
Our evaluation in multi-stage demonstrates the benefit of our approach againist others and HandDiffuse fills the blank in human motion generation.


\section*{Acknowledgments}
We would like to express our sincere gratitude to all the students and faculty members of the VDI Laboratory at ShanghaiTech University for their invaluable assistance and guidance throughout this experiment. We also appreciate the use of the filming equipment provided by the VDI Laboratory. 

Additionally, we would like to acknowledge our collaborators: Sihang Xu, Hongdi Yang, Yiran Liu, Xin Chen, Jingya Wang, Jingyi Yu, and Lan Xu, for their contributions to this work.

\bibliography{aaai25}

\end{document}